\pdfoutput=1
\documentclass[11pt, a4paper, logo]{lumia}

\usepackage[sort&compress]{natbib}
\usepackage{latexsym}
\usepackage{multirow}
\usepackage{array}
\usepackage{xspace}
\usepackage{makecell}
\usepackage{placeins}
\usepackage{wrapfig}

\definecolor{abstractbg}{HTML}{F2F5F8}
\setheadertitle{Don't Blindly Trust It: How Unreliable Feedback Breaks Tool-Using LLM Agents}
\title{Don't Blindly Trust It: How Unreliable Feedback Breaks Tool-Using LLM Agents}
\correspondingemail{\emailicon~\href{mailto:yu_xingrui@a-star.edu.sg}{yu\_xingrui@a-star.edu.sg} \quad $^\ddagger$ Corresponding author.}
\author{%
Chubin Zhang$^1$, Zhenglin Wan$^2$, Xingrui Yu$^{3,4\ddagger}$, Pengfei Zhou$^2$, Wangbo Zhao$^2$, Jingxuan Wu$^5$, Yaxin Zhou$^6$, Ivor Tsang$^{1,3,4}$\\
$^1$ Nanyang Technological University, Singapore\\
$^2$ National University of Singapore, Singapore\\
$^3$ CFAR, Agency for Science, Technology and Research, Singapore\\
$^4$ IHPC, Agency for Science, Technology and Research, Singapore\\
$^5$ Department of Statistics and Operations Research, UNC-Chapel Hill, United States\\
$^6$ Carnegie Mellon University, United States
}

\begin{document}
\begin{abstract}
Tool-augmented agents are typically evaluated by their gains under reliable external feedback. Yet these gains leave open a key counterfactual: when feedback is unreliable, would the agent be better off receiving no task evidence? We study this question with a controlled matched-loop comparison that fixes the agent loop, prompt, action space, and decoding, while varying only the returned observation: faithful, misleading, or absent. Across question answering and fact verification, persistent misleading feedback produces a value inversion: agents that benefit from clean tools can perform worse than the matched no-feedback fallback. On HotpotQA, Qwen2.5-7B reaches 44.8 F1 with clean retrieval and 22.3 F1 with no feedback, but drops to 4.7 F1 under shuffled retrieval. The inversion persists under stronger clean retrieval and locally plausible distractors, but weakens when later clean evidence can repair the trajectory. Early trajectory signals predict many failures, yet simple repairs remain fallback-limited: rejecting bad evidence helps only when the exposed fallback is reliable. These results show that clean-tool gains can overstate tool value, and that matched no-feedback fallback controls are necessary for evaluating tool-augmented agents.
\end{abstract}

\maketitle

\section{Introduction}

Large language model (LLM) agents increasingly solve problems by interacting with external systems rather than generating answers in a single pass. Retrieval, web search, calculators, and execution environments can supply missing facts, intermediate computations, or observable states. These feedback channels are central to why tool use helps: they let the model condition later decisions on information that was not available in the original prompt or in its parametric knowledge \citep{karpas2022mrkl, nakano2021webgpt, yao2023react, schick2023toolformer}. However, the same dependence on external observations also introduces a trust problem. Once an agent updates its trajectory based on tool feedback, success depends not only on whether tools can provide useful information, but also on whether the agent can decide when that information should be trusted, discounted, or ignored.

Prior work has shown that noisy context and misleading external evidence can degrade language-model behavior \citep{cuconasu2024power, wu2024irrelevant, greshake2023not, wang2026agentnoisebench}. However, showing degradation is not the same as measuring the value of tool use. A clean-vs-corrupted comparison tells us whether bad feedback removes some of the benefit of tools, but it does not tell us whether the agent would have been better off without task evidence in the first place. The missing comparison is therefore a fallback counterfactual: under unreliable feedback, does tool use still outperform a matched condition in which the agent receives no task evidence? This distinction is central to our study. ``Corruption hurts'' is expected; ``corruption makes tool use worse than fallback'' means that misleading feedback has inverted the value of the tool.

\begin{wrapfigure}[19]{r}{0.38\textwidth}
    \centering
    \includegraphics[width=0.36\textwidth]{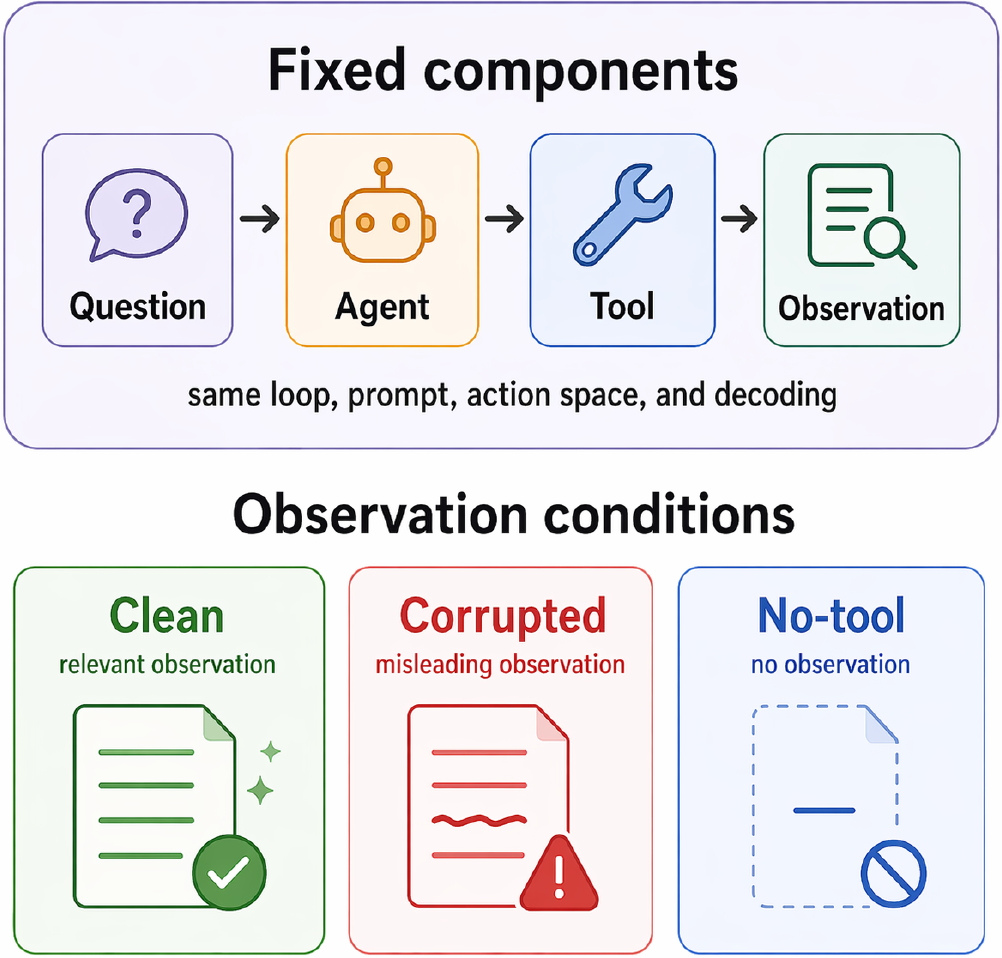}
    \caption{Controlled matched-loop comparison. The agent loop, prompt, action space, and decoding stay fixed; only the returned observation changes: faithful, misleading, or absent.}
    \label{fig:setup}
\end{wrapfigure}

To isolate this counterfactual, we use a controlled matched-loop comparison (Figure~\ref{fig:setup}). Across conditions, the agent follows the same loop, receives the same prompt, has the same action space, and uses the same decoding settings. The only intervention is the observation returned after a tool call. In the clean condition, the environment returns faithful task evidence. In the corrupted condition, it returns misleading evidence through the same interface. In the no-feedback condition, the environment withholds task evidence, so the agent must fall back on its own knowledge while remaining inside the same loop. Thus, no-feedback is a matched fallback condition rather than a separate closed-book prompt, allowing us to attribute performance differences to observation fidelity instead of changes in the agent scaffold.

Under this matched comparison, feedback fidelity changes the value of tool use itself. Reliable observations provide a clear benefit over fallback, but persistent misleading observations can make the same tool-using agent worse than the matched no-feedback condition. On HotpotQA, Qwen2.5-7B reaches 44.8 F1 with clean retrieval and 22.3 F1 with no feedback, but falls to 4.7 F1 under shuffled retrieval. The same qualitative ordering appears on FEVER and on GPT-4o. The fallback baseline is therefore not merely an auxiliary control: it is what distinguishes a reduction in tool benefit from an inversion of tool value.

We then ask whether this inversion is a brittle artifact or a regime-level pattern. Stronger clean retrieval raises the clean ceiling, but corrupted feedback remains below the no-feedback fallback, suggesting that the effect is not created by a weak backend. Replacing shuffled text with locally plausible same-question distractors preserves the inversion, suggesting that it is not limited to synthetic noise. At the same time, the claim has an important boundary: when misleading feedback appears early but later clean evidence can re-enter the trajectory, performance partially recovers. Further checks across model families, a planning-first scaffold, and a calculator-tool pilot suggest that the pattern is not tied to one retriever, one prompt format, or retrieval alone.

Finally, we ask whether the inversion can be detected early or cheaply repaired. This matters because the failure would be less concerning if agents could simply recognize unreliable feedback and ignore it. Early trajectory signals are predictive of later failure, but the failure mode is not uniform across regimes or model scales. Simple repairs also have limited reach. Skeptical prompting provides only modest gains, and re-query verification often hands control back to the agent's own fallback answer. The practical issue is therefore not only whether bad feedback can be detected, but whether the agent has a reliable fallback once that feedback is rejected.

Our contributions are as follows.
\begin{itemize}[leftmargin=1.25em, itemsep=0.2em, topsep=0.2em]
    \item We identify the fallback counterfactual as a missing control in tool-use evaluation, and study it with a matched-loop comparison that isolates observation fidelity across clean, corrupted, and no-feedback conditions.
    \item We show that persistent misleading feedback can invert the value of tool use: agents that benefit from clean tools can perform worse than a matched no-feedback fallback.
    \item We characterize the scope and limits of this inversion: it persists across tasks, model families, scaffolds, and tool types, but recoverable evidence can repair trajectories, early signals only partially diagnose failure, and simple repairs remain constrained by fallback quality.
\end{itemize}
\FloatBarrier

\section{Controlled Matched-Loop Design}
\label{sec:setting}

To study the fallback counterfactual, we need to compare agents that differ in the evidence they receive, not in how they are prompted, decoded, or allowed to act. We therefore use a controlled matched-loop design. Across conditions, the agent follows the same interaction loop, receives the same prompt, uses the same action space and tool interface, and follows the same decoding procedure. The only intervention is the observation returned by the environment after a tool call. We next define the agent setup, feedback conditions, corruption regimes, and evaluation protocol for this comparison.

\subsection{Agent setup and tasks}

Our base setup is a ReAct-style agent \citep{yao2023react} interacting with a Wikipedia-like environment. The agent can search for evidence, inspect returned pages, and finish with an answer. For all feedback conditions, we keep the prompt, tool interface, action space, and interaction budget fixed, so that changes in performance can be attributed to the evidence returned by the environment rather than to changes in the agent scaffold.

We use HotpotQA \citep{yang2018hotpotqa} and FEVER \citep{thorne2018fever} as the main retrieval-based tasks. HotpotQA tests multi-hop question answering, where misleading evidence can redirect a search trajectory across entities or pages. FEVER tests evidence-based fact verification with a smaller output space, providing a complementary setting in which external evidence still shapes the agent's final decision. We evaluate two open-weight model families, Qwen2.5 and Llama-3.1, and include GPT-4o as a stronger proprietary-model check on HotpotQA.

We also include two scope checks beyond the base retrieval agent. First, we evaluate a planning-first variant in which the model writes a short plan before entering the same action loop. This tests whether the inversion is tied to the base ReAct scaffold. Second, we run a calculator-tool pilot on GSM8K \citep{cobbe2021training}. This pilot is not intended as a full additional benchmark branch; instead, it tests whether persistent misleading feedback can also fall below a fallback condition outside retrieval.

\subsection{Feedback conditions}

The central comparison is among three matched observation conditions. In Clean, the environment returns faithful task evidence. In Corrupted, the same tool interface returns misleading evidence. In No Feedback, the environment withholds task evidence, so the agent must fall back on its own knowledge while remaining inside the same loop. Thus, No Feedback is a matched fallback condition rather than a separate closed-book prompt.

This comparison isolates the role of observation fidelity. The gap between Clean and No Feedback measures how much reliable external evidence helps under the fixed agent loop. The gap between No Feedback and Corrupted measures whether misleading evidence merely reduces that benefit or instead makes tool use worse than fallback. The latter case is the value inversion studied in this paper.

\subsection{Corruption and recovery regimes}

We construct corrupted observations to separate persistent misleading feedback from recoverable evidence. In Shuffled, observations are replaced with paragraphs drawn from other question-local pages rather than the requested evidence. This creates a persistent-misleading regime in which the agent repeatedly receives off-target evidence. In Plausible, retrieval is restricted to same-question distractors, producing topic-adjacent but incorrect evidence. This condition tests whether the effect survives realistic retrieval mismatch rather than only synthetic noise.

We also include a recoverable-evidence regime, Conflicting. In this setting, the agent receives an early misleading observation, but later calls can again return clean evidence. This condition is important because our claim is not that every wrong observation is equally harmful. Rather, the central risk arises when misleading feedback persistently redirects the trajectory and prevents the agent from recovering.

As a retrieval-strength check, we compare the default HotpotQA retrieval backend with a stronger oracle-style backend that can return gold supporting pages in the clean condition. This check asks whether the main inversion is caused by a weak clean retriever rather than by over-trust in misleading feedback. Full implementation details are given in Appendix~\ref{app:implementation}.

\subsection{Evaluation}

We report standard task metrics for the main results: exact match and token-level F1 for HotpotQA, label accuracy for FEVER, and exact numeric accuracy for GSM8K. For analyses that require paired trajectory outcomes, we additionally use a binary success indicator based on normalized answer overlap. This indicator marks a trajectory as successful when answer overlap reaches F1 $\ge 0.6$, and is used only for trajectory-level analyses and paired significance tests. Unless otherwise noted, paired comparisons use exact McNemar tests, and confidence intervals for headline metric gaps are estimated with paired bootstrap resampling.

\section{Value Inversion Under Misleading Feedback}
\label{sec:main}

We now establish the paper's central empirical finding. Under a matched-loop comparison, misleading feedback does not merely reduce the benefit of tool use. In persistent-misleading regimes, it can make the same tool-using agent perform worse than the matched no-feedback fallback.

\subsection{Core matched comparison}

We begin with the comparison that motivates the paper: Clean, No Feedback, and Shuffled, our default corrupted condition. The goal is not simply to show that corruption hurts. Rather, the comparison asks whether tool use remains preferable to falling back without task evidence. If shuffled feedback falls below the no-feedback fallback, then misleading feedback has not merely reduced the benefit of the tool; it has inverted the value of using it.

\begin{figure}[!htbp]
    \centering
    \includegraphics[width=0.94\textwidth]{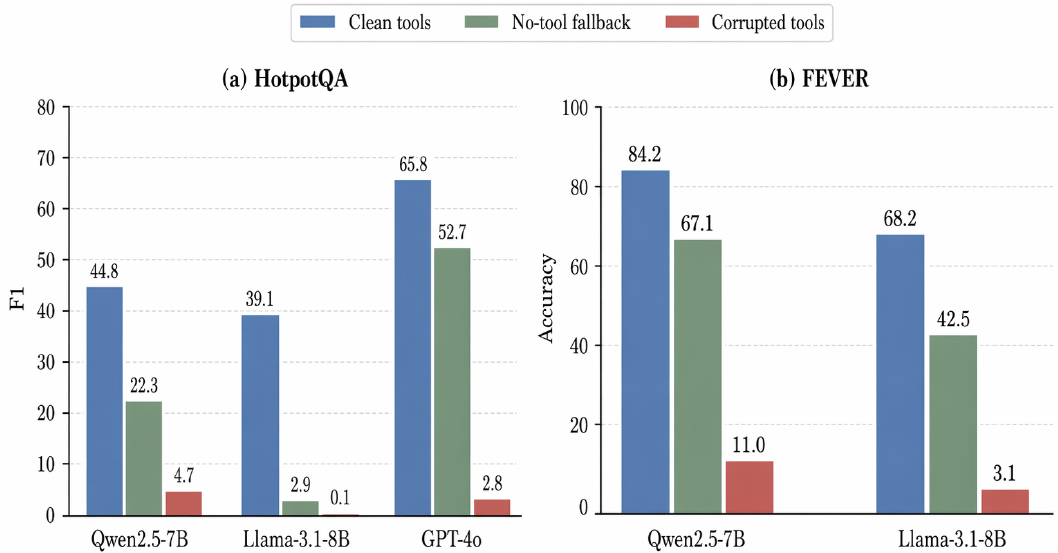}
    \caption{Core matched comparison under standard task metrics. Across HotpotQA and FEVER, reliable observations improve over the no-feedback fallback, while shuffled misleading feedback falls below it. GPT-4o is shown for HotpotQA only.}
    \label{fig:main_results}
\end{figure}

Figure~\ref{fig:main_results} and Table~\ref{tab:main} show a consistent ordering: reliable observations help, but persistent misleading observations are worse than receiving no task evidence. This pattern means that clean-tool gains alone do not determine the value of tool use. The same dependence on external observations that helps under reliable feedback can become harmful when those observations are misleading.

The effect is especially clear on HotpotQA. Qwen2.5-7B benefits substantially from clean retrieval, yet drops far below the matched fallback under shuffled feedback. Paired analyses support the same interpretation: for this model, the clean condition exceeds No Feedback by 22.5 F1 points, while No Feedback exceeds Shuffled by 17.6 F1 points under paired bootstrap estimates; exact paired tests on trajectory success support the same ordering. The same qualitative pattern appears across the other tested HotpotQA models and on FEVER. Thus, the value inversion is not limited to a single model family, nor to free-form multi-hop question answering. It also appears in evidence-based fact verification, where the output space is smaller and the metric is exact label accuracy.

\begin{table}[t]
\centering
\small
\setlength{\tabcolsep}{6pt}
\caption{Standard-metric results for the core matched comparison. The consistent ordering, Clean $>$ No Feedback $>$ Shuffled, shows that reliable observations help while persistent misleading observations can fall below the matched fallback. HotpotQA uses 300 examples for open-weight models and 100 examples for GPT-4o; FEVER uses 292 examples.}
\label{tab:main}
\begin{tabular}{llccc}
\toprule
Task & Model & Clean & No Feedback & Shuffled \\
\midrule
\multirow{4}{*}{HotpotQA (F1)}
& Qwen2.5-3B & 42.1 & 11.4 & 0.3 \\
& Qwen2.5-7B & 44.8 & 22.3 & 4.7 \\
& Llama-3.1-8B & 39.1 & 2.9 & 0.1 \\
& GPT-4o & 65.8 & 52.7 & 2.8 \\
\midrule
\multirow{2}{*}{FEVER (Accuracy)}
& Qwen2.5-7B & 84.2 & 67.1 & 11.0 \\
& Llama-3.1-8B & 68.2 & 42.5 & 3.1 \\
\bottomrule
\end{tabular}
\end{table}
\FloatBarrier

\subsection{Retrieval strength does not explain the inversion}

A natural concern is that the HotpotQA result might be driven by a weak retrieval backend. If the clean retriever is poor, then the apparent value of tool use could be underestimated, and the comparison might reflect retrieval quality rather than over-trust in misleading feedback. To test this explanation, we use an oracle-style retrieval intervention that can return gold supporting pages in the clean condition while preserving the same agent loop and interface.

The intervention produces the expected effect on clean retrieval: it raises the clean ceiling substantially. However, it does not change the main comparison. The no-feedback fallback remains at roughly the same level, and shuffled feedback remains far below it. Thus, stronger retrieval increases the value of faithful observations, but it does not remove the value inversion under misleading observations.

This separation is important. A weak retriever can suppress how much clean tools appear to help, but it cannot explain why misleading feedback makes the agent worse than fallback. The core harm pattern therefore comes from the agent's dependence on misleading observations, not merely from a weak clean retrieval backend.

\subsection{Persistent versus recoverable misleading evidence}

Shuffled feedback is a useful stress test, but it is deliberately severe. It therefore leaves open two questions. First, does the inversion persist when the wrong evidence is locally plausible rather than obviously off-target? Second, does any early misleading observation cause the same failure, or can later clean evidence repair the trajectory? We use Plausible and Conflicting to separate these cases.

\begin{wraptable}[17]{r}{0.43\textwidth}
\centering
\small
\setlength{\tabcolsep}{4pt}
\caption{Retrieval-strength and corruption-regime checks on Qwen2.5-7B HotpotQA. Oracle retrieval raises the clean ceiling, while Shuffled and Plausible remain below fallback and Conflicting recovers above it.}
\label{tab:retrieval}
\begin{tabular}{lcc}
\toprule
Setting (Qwen2.5-7B) & EM & F1 \\
\midrule
heuristic Clean & 33.7 & 44.8 \\
heuristic No Feedback & 15.3 & 22.3 \\
heuristic Shuffled & 2.0 & 4.7 \\
oracle Clean & 42.0 & 54.9 \\
oracle No Feedback & 16.3 & 22.6 \\
oracle Shuffled & 2.3 & 4.8 \\
Plausible & 6.7 & 9.5 \\
Conflicting & 29.7 & 39.1 \\
\bottomrule
\end{tabular}
\end{wraptable}

\paragraph{Plausible same-question distractors.}
In Plausible, the agent receives distractor evidence drawn from the same HotpotQA question packet. The evidence is wrong, but it is topic-adjacent and often entity-adjacent, making it closer to a realistic retrieval mismatch than shuffled paragraph replacements. As Table~\ref{tab:retrieval} shows, Plausible remains well below the no-feedback fallback. Thus, the inversion is not limited to obviously irrelevant noise: locally plausible wrong evidence can still be worse than receiving no task evidence.

\paragraph{Recoverable evidence as a boundary.}
Conflicting tests a different case. The agent receives an early misleading observation, but later calls can again return clean evidence. In this regime, performance recovers above the no-feedback fallback, although it remains below the clean condition. We therefore treat Conflicting as a recoverable-evidence regime rather than as a counterexample to the main claim. Its role is to identify a boundary: early corruption is not destiny when clean evidence can re-enter and repair the trajectory.

\paragraph{Summary.}
Taken together, these results support a narrow but strong claim: in persistent-misleading regimes, corrupted feedback can make tool use worse than a matched no-feedback fallback. This claim holds under standard task metrics, stronger clean retrieval, locally plausible distractors, two open-weight model families, and GPT-4o. At the same time, the claim is not unconditional. Conflicting identifies a recoverable-evidence regime in which later clean observations restore much of the lost performance. This boundary is important for the rest of the paper: the failure is not caused by any single wrong observation, but by misleading feedback that persistently redirects the agent's trajectory.

\section{Scope Beyond Base Retrieval}
\label{sec:scope}

The previous section establishes the value inversion in the base matched-loop retrieval setting. We next ask whether the pattern survives natural changes to the setup. These checks are not meant to establish a universal law of tool use; rather, they test whether the inversion is tied to one model family, one agent scaffold, or retrieval alone.

Table~\ref{tab:generalization} summarizes the scope checks. GPT-4o follows the same HotpotQA ordering as the open-weight models, suggesting that stronger model capability does not by itself eliminate value inversion. A planning-first scaffold also preserves the pattern: planning can shift the clean/fallback tradeoff, but corrupted feedback still falls below fallback. Finally, a GSM8K calculator-tool pilot shows the same qualitative ordering outside retrieval, where the external feedback is symbolic rather than textual. These results suggest that the inversion is not specific to the base ReAct retrieval setup, while remaining deliberately limited in scope.

\begin{center}
\begin{minipage}{0.72\textwidth}
\captionsetup{type=table,width=\linewidth}
\caption{Scope checks beyond the base ReAct retrieval setup. The ordering persists for GPT-4o, planning-first agents, and a calculator-tool pilot. Metrics are F1 for HotpotQA and accuracy for GSM8K.}
\label{tab:generalization}
\centering
\small
\setlength{\tabcolsep}{4pt}
\begin{tabular}{lccc}
\toprule
Setting & Clean & Fallback & Corrupted \\
\midrule
GPT-4o HotpotQA & 65.8 & 52.7 & 2.8 \\
Qwen2.5-7B plan-then-act & 47.8 & 16.5 & 0.2 \\
Llama-3.1-8B plan-then-act & 40.0 & 0.7 & 0.0 \\
Qwen2.5-7B GSM8K+calculator & 83.0 & 52.0 & 14.0 \\
\bottomrule
\end{tabular}
\end{minipage}
\end{center}

\section{Diagnosing the Failure Structure}
\label{sec:diagnosis}

Having established the value inversion, we next ask whether the failure has observable structure before the final answer. This matters because a failure visible only in the final metric is difficult to monitor or repair during the agent loop. We do not claim a single causal mechanism. Instead, we focus on diagnostic evidence: early trajectory signals predict later failure, while the form of failure varies across corruption regimes and models.

\subsection{Early trajectory signals predict later failure}

We first test whether failure can be predicted from information available after the first tool interaction. We train a lightweight logistic regression using only first-step observable features, including lexical overlap between the question, query, returned title, and observation; length features; and simple uncertainty markers in the first-step text. Gold support metadata is used only for descriptive analysis, not as model input.

Table~\ref{tab:e6} shows that these early signals are predictive on held-out grouped evaluation for both open-weight model families. Importantly, the result is not driven only by the easiest Shuffled cases. When evaluation is restricted to the cleaner but harder Clean+Plausible subset, the predictor remains above chance. Thus, the failure is not merely a final-answer artifact: the first tool interaction already exposes signals that separate many later successes from failures.

\begin{center}
\begin{minipage}{0.62\textwidth}
\captionsetup{type=table,width=\linewidth}
\caption{Held-out first-step failure prediction. The main setting uses Clean, Shuffled, and Plausible trajectories; the stricter subset removes Shuffled examples.}
\label{tab:e6}
\centering
\small
\setlength{\tabcolsep}{4pt}
\begin{tabular}{lcc}
\toprule
Model & Main AUC & Clean+Plausible AUC \\
\midrule
Qwen2.5-7B & 0.796 $\pm$ 0.036 & 0.751 $\pm$ 0.043 \\
Llama-3.1-8B & 0.849 $\pm$ 0.023 & 0.763 $\pm$ 0.035 \\
\bottomrule
\end{tabular}
\end{minipage}
\end{center}

\subsection{Failure can be concentrated or distributed}

Predictability at the first step does not imply that all models fail in the same way. To separate different failure patterns, we compare targeted corruption with full corruption using the trajectory-level success heuristic.

\begin{wrapfigure}[20]{r}{0.47\textwidth}
    \centering
    \includegraphics[width=0.45\textwidth]{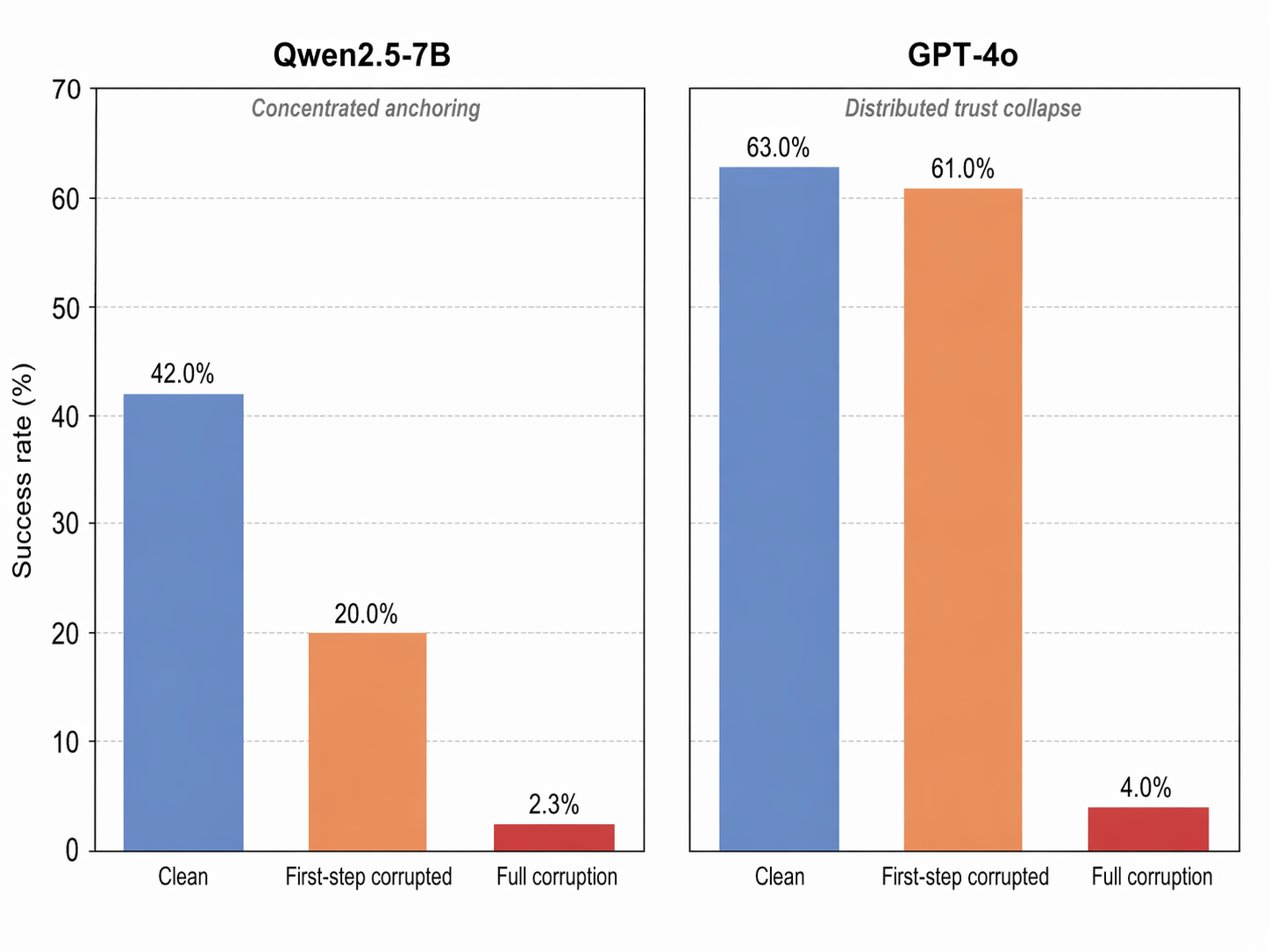}
    \caption{Failure structure under targeted corruption. Qwen2.5-7B shows concentrated anchoring, where first-step corruption accounts for a large share of the damage. GPT-4o is more robust to a single corrupted observation, but still collapses under persistent corruption.}
    \label{fig:failure_topology}
\end{wrapfigure}

Figure~\ref{fig:failure_topology} shows two qualitatively different patterns. For Qwen2.5-7B, corrupting only the first search result accounts for a large share of the total damage. We describe this as \emph{concentrated anchoring}: an early misleading observation can capture the trajectory before later evidence has a chance to repair it. GPT-4o shows a different pattern. It is much more robust to a single corrupted first observation, but still collapses under persistent corruption. We describe this as \emph{distributed trust collapse}: no single early observation is decisive, yet cumulative misleading feedback eventually overwhelms the trajectory. These terms are descriptive labels for the observed failure patterns, not claims of a single underlying mechanism.

This distinction helps explain why the same value inversion can persist across model scales without implying a universal failure mode. Smaller open-weight models may be more vulnerable to early anchoring, while stronger models may absorb isolated errors but still fail when misleading feedback persists.

\subsection{Recoverable evidence breaks first-step determinism}

The Conflicting condition provides a useful boundary case. In this regime, the agent receives an early misleading observation, but later clean evidence can re-enter the trajectory. If first-step evidence fully determined the outcome, the first-step predictor should transfer cleanly to this setting. It does not: when the predictor trained on Clean, Shuffled, and Plausible trajectories is evaluated on Conflicting trajectories for Qwen2.5-7B, AUC drops to 0.516.

This near-chance transfer is informative rather than problematic. It shows that early bad evidence is not destiny once later clean evidence can repair the trajectory. Therefore, Conflicting bounds the scope of the first-step diagnostic story, just as it bounds the main value-inversion claim. The central risk is persistent misleading feedback, not any isolated wrong observation.

\subsection{Hedging is not skepticism}

We also observe that verbal uncertainty is not the same as behavioral skepticism. Under harmful conditions, the model often becomes more verbose and more visibly uncertain. However, on Qwen2.5-7B, higher first-step hedging counts and longer first-step text are associated with failure rather than recovery. This pattern is consistent with recent observations that confidence signals can diverge across evidence and verification tools \citep{xuan2026confidence}.

In other words, the agent can sound doubtful while still following misleading evidence. This is why we distinguish surface-level hedging from actual evidence reweighting. A useful fallback policy must do more than express uncertainty; it must change how the agent acts when observations are unreliable.

\section{Simple Repairs Are Often Fallback-Limited}
\label{sec:mitigation}

A natural response to value inversion is to ask whether simple repairs can make agents ignore unreliable feedback. This question matters because the failure would be less concerning if it could be removed by a small prompt change or a lightweight verification step. We therefore evaluate two deliberately minimal interventions: skeptical prompting, which explicitly warns the agent that tool feedback may be unreliable, and re-query verification, which accepts an observation only when repeated queries return sufficiently consistent outputs. Details are given in Appendix~\ref{app:verification-details}.

These repairs help only partially. Skeptical prompting yields modest gains, but the improvement is small relative to the damage caused by persistent misleading feedback. Re-query verification is more informative. It can help on Qwen2.5-7B, but it is not uniformly beneficial and is harmful on Llama-3.1-8B in our experiments (Appendix Table~\ref{tab:mitigation}). Thus, the issue is not simply that the model needs to be reminded to distrust tools, nor that noisy observations can always be removed by a lightweight consistency check.

The reason is that verification changes the problem rather than solving it. When the verifier rejects a suspicious observation, the agent is effectively returned to its fallback policy. On Qwen2.5-7B at high corruption, the verifier rejects nearly all corrupted observations and has no measured false accepts, yet the downstream gain remains limited. This suggests that detector quality is not the only bottleneck. Even a strong filter can help only when the agent has a reliable fallback once the observation is removed.

A same-model verification analysis supports this fallback-limited view. Across several proxies for parametric competence, examples with stronger fallback support tend to benefit from verification, while examples with weaker fallback support can be harmed. The cleanest oracle-backed split shows the expected sign pattern, but the full pre-registered threshold is not consistently met. We therefore treat this as directional evidence rather than a formal proof: verification appears most useful when the agent can answer well after unreliable evidence has been rejected.

The broader implication is that filtering and fallback must be evaluated together. Clean tools help when feedback is reliable, but coupling an agent to external observations makes feedback fidelity a first-class control variable. Under persistent misleading feedback, tool use can be worse than the matched no-feedback fallback. When simple repairs reject that feedback, their benefit is bounded by the quality of the fallback they expose.

\section{Related Work}

Prior work has shown that language models are sensitive to unreliable external evidence. Retrieval-augmented models combine parametric generation with retrieved external evidence \citep{lewis2020retrieval, guu2020realm}. In such settings, noisy or low-quality retrieval can distort downstream answers, and even superficially irrelevant context can change model behavior \citep{shi2023distracted, cuconasu2024power, wu2024irrelevant}. Work on knowledge conflict similarly shows that models can accept coherent external evidence even when it conflicts with parametric memory \citep{xie2023adaptive}. Related studies examine noisy or counterfactual retrieval in RAG \citep{chen2023rgb}, cases where retrieval can hurt and filtering can help \citep{mallen2023when, yoran2024making, yu2023chainnote}, and imperfect tool environments for agents, including noisy tool outputs, distractor-rich contexts, and adversarial web content \citep{wang2026agentnoisebench, kwak2025toolhaystack, shah2026syntheticweb}. Our work is closest to this line, but asks a different comparison question. Rather than measuring robustness degradation alone, we center the fallback counterfactual: under a matched agent loop, does corrupted feedback remain better than receiving no task evidence? This comparison distinguishes a reduction in clean-tool benefit from an inversion of tool value.

A second related line studies whether models can critique, verify, or correct their own reasoning. Prior work finds that self-correction is often unreliable, and that models are generally better at fixing an identified mistake than at detecting one in the first place \citep{huang2023cannot, tyen2023cannot}. This is relevant to our skeptical-prompting and re-query verification results. Security work on prompt injection and instruction hierarchy also emphasizes that tool outputs should be treated as untrusted inputs rather than automatically followed \citep{greshake2023not, yi2023benchmarking, zhan2024injecagent, wallace2024instruction}. Our setting is narrower: we study epistemic over-trust in corrupted observations, including cases without explicit adversarial instructions. The mitigation question is also different. It is not only whether suspicious evidence can be filtered, but what the agent can do after that evidence is rejected. This is why our analysis focuses on fallback-limited repair: filtering unreliable observations helps only when the exposed fallback policy is strong enough.

\section{Conclusion}

We study a simple but under-specified question in tool-augmented LLM agents: when feedback becomes unreliable, does tool use remain preferable to falling back without task evidence? Using a controlled matched-loop comparison that fixes the agent scaffold and varies only observation fidelity, we find that clean feedback remains strongly beneficial, but persistent misleading feedback can invert the value of tool use, driving performance below a matched no-feedback fallback.

This inversion is not explained away by weak retrieval or synthetic noise. Stronger clean retrieval raises the clean ceiling, while locally plausible distractors still fall below fallback. At the same time, the phenomenon has an important boundary: when later clean evidence can re-enter the trajectory, performance can recover. This boundary suggests that the main risk is not an isolated wrong observation, but persistent misleading feedback that redirects the agent's trajectory.

Early trajectory signals can predict later failure, but failure structure differs across models and regimes. Simple repairs are often fallback-limited: filtering suspicious observations helps only when the agent has a reliable fallback after that evidence is rejected. Thus, the problem is not only how to detect unreliable feedback, but what the agent should do once feedback can no longer be trusted.

These results caution against evaluating tool use only through clean-case gains. Clean tools show what external feedback can add when reliable; the fallback counterfactual shows whether tool use remains valuable when feedback is not. Matched no-feedback fallback should therefore be treated as a first-class control in evaluating tool-augmented agents.

\section{Limitations}

Our study prioritizes controlled comparison over benchmark breadth, which leaves several limitations. First, the proprietary-model analysis covers only one provider and is limited to HotpotQA. This is sufficient to test whether the inversion disappears for a stronger model, but not to establish a broad scaling trend across frontier systems. Second, the calculator experiment is a pilot rather than a full cross-tool benchmark branch. It shows that the pattern can appear outside retrieval, but more tool types are needed before making stronger claims about tool use in general.

Third, our verification analysis provides directional evidence for fallback-limited repair rather than a definitive within-model theory. The same-model splits are consistent with the view that verification helps when fallback competence is strong and can hurt when fallback competence is weak, but the pre-registered threshold is not consistently met. Finally, the retrieval-strength analysis compares a lightweight default backend with an oracle-style backend, but does not include intermediate retrieval systems such as BM25 or dense retrievers. These choices reflect the paper's main goal: isolating the fallback counterfactual under matched loops, rather than exhaustively benchmarking every model, retriever, or tool setting.

\bibliographystyle{plainnat}
\bibliography{project_refs,custom}

\clearpage
\appendix
\begin{center}
{\Large\bfseries Appendix}
\end{center}

\section{Implementation Details}
\label{app:implementation}

This appendix gives implementation details for the matched-loop experiments. Across feedback conditions, the prompt, action space, interaction budget, and decoding procedure are fixed; only the returned observation changes.

\subsection{Agent loop, decoding, and sampling}
\label{app:agent-details}

For HotpotQA and FEVER, the agent follows a ReAct-style loop with search, lookup, and finish actions. The interaction budget is 8 steps. Local models are decoded with temperature 0.0, a maximum of 400 generation tokens per step, deterministic seed 42, and stop sequences keyed to the next observation field. GPT-4o uses the same temperature and stop rules through a batched chat-completions wrapper.

Most local HotpotQA experiments use a 300-example seed-42 sample from the distractor development split. FEVER uses the 292-example filtered validation subset described in the main text. GPT-4o and GSM8K experiments use 100 examples and are used as scope checks rather than full benchmark branches.

\subsection{Retrieval backends}
\label{app:retrieval-details}

The default HotpotQA retriever is intentionally lightweight. It first attempts an exact title match, then a fuzzy title match using \texttt{SequenceMatcher} with threshold 0.3, and finally lexical-overlap search over the available titles and paragraphs. The oracle-style backend keeps the same search interface but restricts candidates to gold supporting titles. If the default matching sequence returns no candidate, the oracle backend emits a deterministic supporting page.

This oracle-style backend is used only as a retrieval-strength check. Its purpose is to raise the clean retrieval ceiling without changing the agent loop, prompt, action space, or decoding procedure.

\subsection{Corruption construction}
\label{app:corruption-details}

For HotpotQA, all corrupted retrieval conditions are implemented at the observation level. In Shuffled, returned observations are replaced with off-target paragraphs from the question context rather than the requested evidence. In Plausible, search is restricted to same-question distractor titles, namely pages in the HotpotQA context packet that are not listed in the supporting facts. In Conflicting, the first search call is forced onto a same-question distractor, after which the clean backend resumes.

For GSM8K, calculator corruption is persistent per expression. Integer results are shifted by roughly 10\%, with a minimum offset of 1. Decimal results are shifted by at least 0.5. The same corrupted value is cached for repeated queries of the same expression.

\subsection{GPT-4o details}
\label{app:gpt4o-details}

GPT-4o experiments use \texttt{gpt-4o-2024-08-06} through the OpenAI chat-completions API. Each HotpotQA run uses the same 100 seed-42 sampled questions, and targeted-corruption runs reuse the same seed. Under Shuffled feedback, GPT-4o takes 7.81 steps on average, close to the 8-step cap. This supports the interpretation that the model is floundering under misleading evidence rather than failing trivially at the parser level.

\section{Failure-Prediction Details}
\label{app:prediction-details}

\subsection{Feature construction and evaluation}
\label{app:e6-details}

The first-step predictor uses only features observable after the first tool interaction. The features fall into four groups: lexical overlaps among the question, search query, first returned title, and first observation; length features such as query length, first-thought length, and observation length; hedging and uncertainty markers in the first-step text; and lightweight surface cues such as years and punctuation density. Gold support-title metadata is used only for descriptive analysis and is never included as a predictor.

The predictor is a standardized logistic regression with class-balanced training and default L2 regularization. We evaluate it with held-out 5-fold cross-validation grouped by \texttt{question\_id}. All reported AUCs are threshold-free and use held-out probabilities rather than a tuned decision threshold.

\subsection{Cross-validation results}

On the main setting, which combines Clean, Shuffled, and Plausible trajectories, Qwen2.5-7B obtains per-fold AUCs of [0.835, 0.774, 0.737, 0.785, 0.850], yielding 0.796$\pm$0.036 overall. Llama-3.1-8B obtains 0.849$\pm$0.023. On the stricter Clean+Plausible subset, which removes the easiest Shuffled examples, AUC remains 0.751$\pm$0.043 on Qwen2.5-7B and 0.763$\pm$0.035 on Llama-3.1-8B.

\subsection{Feature trends and transfer}

Across both model families, larger title--query and title--question lexical overlaps are associated with later success. Longer first thoughts, longer first search queries, and more hedging markers are associated with later failure. This supports the main-text observation that verbal uncertainty does not necessarily translate into behavioral skepticism.

When the Qwen2.5-7B predictor trained on Clean, Shuffled, and Plausible trajectories is evaluated on Conflicting trajectories, AUC drops to 0.516. Predicted failure at threshold 0.5 is 71.7\%, compared with an actual failure rate of 62.3\%. This near-chance transfer supports the interpretation of Conflicting as a recoverable-evidence boundary regime: when later clean evidence can repair the trajectory, first-step signals no longer rank outcomes reliably.

\section{Verification and Repair Details}
\label{app:verification-details}

\subsection{Re-query verifier}

The re-query verifier is environment-side and not learned. For each query, the environment samples two candidate observations independently and computes Jaccard overlap over token sets. If the overlap is at least 0.5, the observation is accepted; otherwise, it is replaced with a no-results message. False accepts are measured against the simulator's corruption labels, not against downstream answer correctness.

Across corruption rates on Qwen2.5-7B, the verifier accepts $527/1043$ observations at 30\% corruption, $257/1106$ at 50\%, and $106/1279$ at 70\%. It has zero measured false accepts under this simulator-label criterion in all three settings.

\subsection{Mitigation results}

Table~\ref{tab:mitigation} reports HotpotQA F1 deltas for skeptical prompting and re-query verification across corruption rates. Skeptical prompting helps modestly across both model families. Verification is more model-dependent: it helps Qwen2.5-7B, but is harmful for Llama-3.1-8B in these runs.

\begin{table}[t]
\centering
\small
\setlength{\tabcolsep}{4pt}
\caption{Mitigation deltas in HotpotQA F1. Skeptical prompting helps only modestly, while verification is strongly model-dependent.}
\label{tab:mitigation}
\begin{tabular}{llccc}
\toprule
Method & Model & 30\% & 50\% & 70\% \\
\midrule
skeptical prompt & Qwen & +4.9 & +2.9 & +5.7 \\
skeptical prompt & Llama & +3.2 & +4.6 & +1.4 \\
verification & Qwen & +0.5 & +3.6 & +5.9 \\
verification & Llama & -1.7 & -3.3 & -4.8 \\
\bottomrule
\end{tabular}
\end{table}

\subsection{Within-model verification analysis}
\label{app:e5-details}

To probe whether verification is fallback-limited, we split the same 300 HotpotQA questions by several proxies for parametric competence on Qwen2.5-7B. The first split uses whether the model is correct in the No Feedback condition. The second uses direct closed-book QA correctness. The third is an ensemble-strict split: examples are easy if at least 2 of 3 closed-book prompts are correct, hard if 0 of 3 are correct, and examples at exactly 1 of 3 are excluded as ambiguous. We use this split, together with its oracle-retrieval variant, because it gives the cleanest separation for the main-text analysis.

\begin{table}[!ht]
\centering
\small
\setlength{\tabcolsep}{4pt}
\caption{Within-model verification deltas on Qwen2.5-7B across four parametric-strength proxies. Signs are directionally consistent across proxies: verification helps examples with stronger parametric support and is neutral-to-harmful on examples with weaker parametric support. The hard-side effect reaches $p=0.03$ at 50\% corruption under the oracle + ensemble-strict proxy, but the pre-registered full threshold ($p<0.01$ on both sides) is not uniformly met. The main text therefore reports this result as directional evidence consistent with a fallback-limited tradeoff.}
\label{tab:e5-full}
\begin{tabular}{lrrrr}
\toprule
Corruption & Easy $\Delta$ & Easy $p$ & Hard $\Delta$ & Hard $p$ \\
\midrule
\multicolumn{5}{l}{\textbf{No Feedback proxy (easy=60, hard=240)}} \\
30\% & +11.7 & 0.167 & -2.5 & 0.497 \\
50\% & +26.7 & 0.004 & -2.9 & 0.410 \\
70\% & +50.0 & $<$0.001 & -5.4 & 0.053 \\
\midrule
\multicolumn{5}{l}{\textbf{Closed-book proxy (easy=73, hard=227)}} \\
30\% & +1.4 & 1.000 & +0.0 & 1.000 \\
50\% & +20.5 & 0.014 & -2.6 & 0.471 \\
70\% & +27.4 & $<$0.001 & -1.3 & 0.755 \\
\midrule
\multicolumn{5}{l}{\textbf{Ensemble-strict proxy (easy=72, hard=220)}} \\
30\% & +1.4 & 1.000 & -0.5 & 1.000 \\
50\% & +20.8 & 0.011 & -3.6 & 0.291 \\
70\% & +27.8 & $<$0.001 & -2.7 & 0.418 \\
\midrule
\multicolumn{5}{l}{\textbf{Oracle retrieval + ensemble-strict}} \\
30\% & +12.5 & 0.093 & -5.0 & 0.161 \\
50\% & +9.7 & 0.281 & -8.2 & 0.030 \\
70\% & +26.4 & 0.001 & -5.0 & 0.169 \\
\bottomrule
\end{tabular}
\end{table}

\end{document}